\documentclass[10pt,twocolumn,letterpaper]{article}

\usepackage{cvpr}
\usepackage{times}
\usepackage{epsfig}
\usepackage{graphicx}
\usepackage{amsmath}
\usepackage{amssymb}

\usepackage{booktabs}

\usepackage[pagebackref=true,breaklinks=true,letterpaper=true,colorlinks,bookmarks=false]{hyperref}

\cvprfinalcopy 

\def\cvprPaperID{****} 

\ifcvprfinal\fi
\begin{document}

\title{ConTNet: Why not use convolution and transformer at the same time?}

\author{Haotian Yan\textsuperscript{1,2}\thanks{contributed equally}  \quad Zhe Li\textsuperscript{2}\footnotemark[1] \quad Weijian Li\textsuperscript{2} \quad Changhu Wang\textsuperscript{2} \quad Ming Wu\textsuperscript{1} \quad Chuang Zhang\textsuperscript{1} \\
\textsuperscript{1}School of AI, Beijing University of Posts and Telecommunications. \\ {\tt\small \{yanhaotian, wuming, zhangchuang\}@bupt.edu.cn}\\
\textsuperscript{2}ByteDance AI Lab, Beijing. \\ {\tt\small \{lizhe.axel, liweijian, wangchanghu\}@bytedance.com}

}

\maketitle

\begin{abstract}
   Although convolutional networks (ConvNets) have enjoyed great success in computer vision (CV), it suffers from capturing global information crucial to dense prediction tasks such as object detection and segmentation. In this work, we innovatively propose ConTNet (Convolution-Transformer Network), combining transformer with ConvNet architectures to provide large receptive fields. Unlike the recently-proposed transformer-based models (e.g., ViT, DeiT) that are sensitive to hyper-parameters and extremely dependent on a pile of data augmentations when trained from scratch on a midsize dataset (e.g., ImageNet1k), ConTNet~\cite{yan2021contnet} can be optimized like normal ConvNets (e.g., ResNet) and preserve an outstanding robustness. It is also worth pointing that, given identical strong data augmentations, the performance improvement of ConTNet is more remarkable than that of ResNet. We present its superiority and effectiveness on image classification and downstream tasks. For example, our ConTNet achieves $81.8\%$ top-1 accuracy on ImageNet which is the same as DeiT-B with less than $40\%$ computational complexity. ConTNet-M also outperforms ResNet50 as the backbone of both Faster-RCNN (by $2.6\%$) and Mask-RCNN (by $3.2 \%$) on COCO2017 dataset. We hope that ConTNet could serve as a useful backbone for CV tasks and bring new ideas for model design. The code will be released at \href{https://github.com/yan-hao-tian/ConTNet}{https://github.com/yan-hao-tian/ConTNet}.
\end{abstract}

\section{Introduction}

\begin{figure}[t]
\begin{center}
   \includegraphics[width=0.4\textwidth]{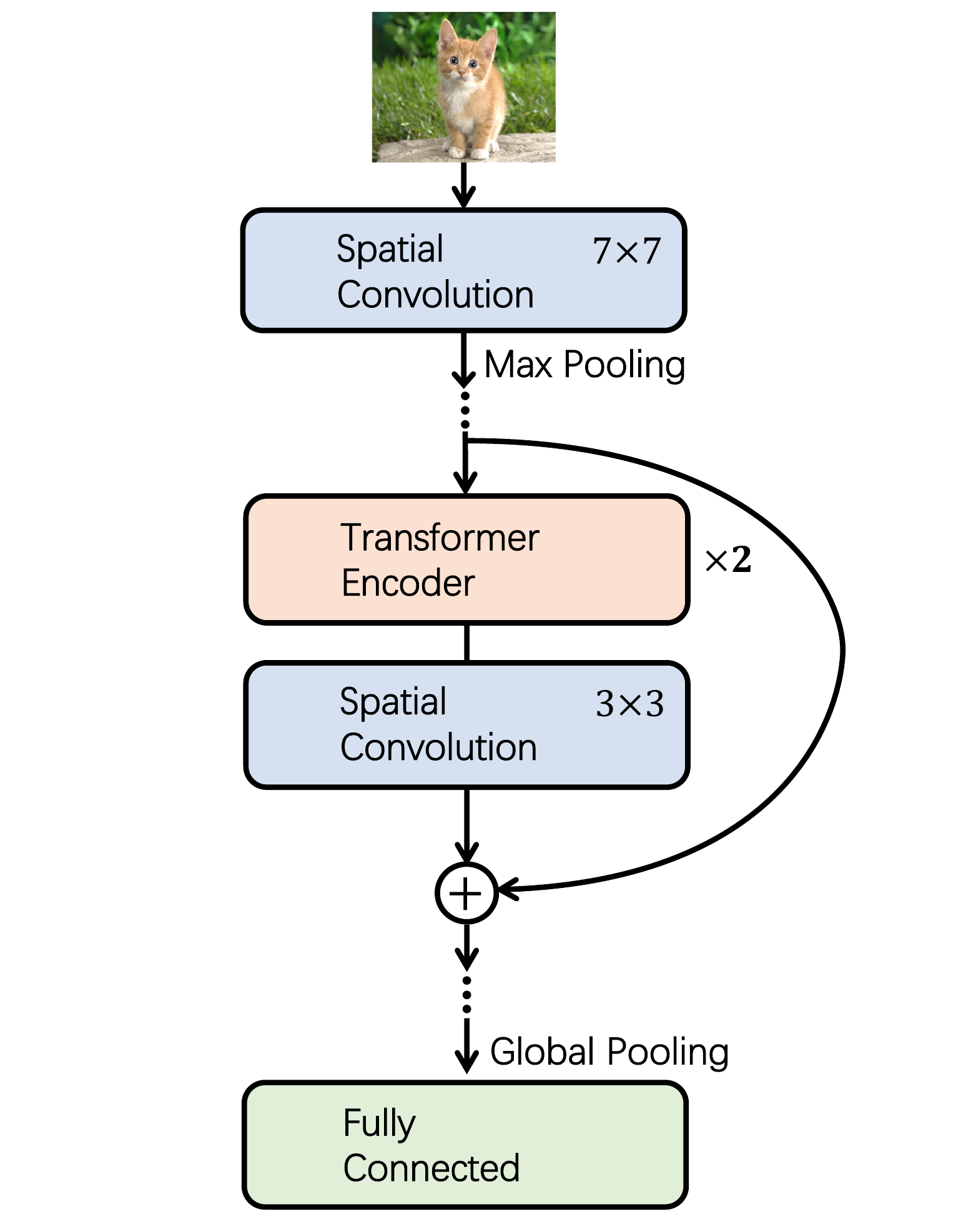}
\end{center}
   \caption{Illustration of the proposed ConTNet framework. ConTNet contains multiple ConT blocks, which are composed of two transformer encoder layers and a convolution layer.}
\label{fig: ConT block}
\end{figure}

\begin{figure*}[t]
\begin{center}
   \includegraphics[width=1.0\textwidth]{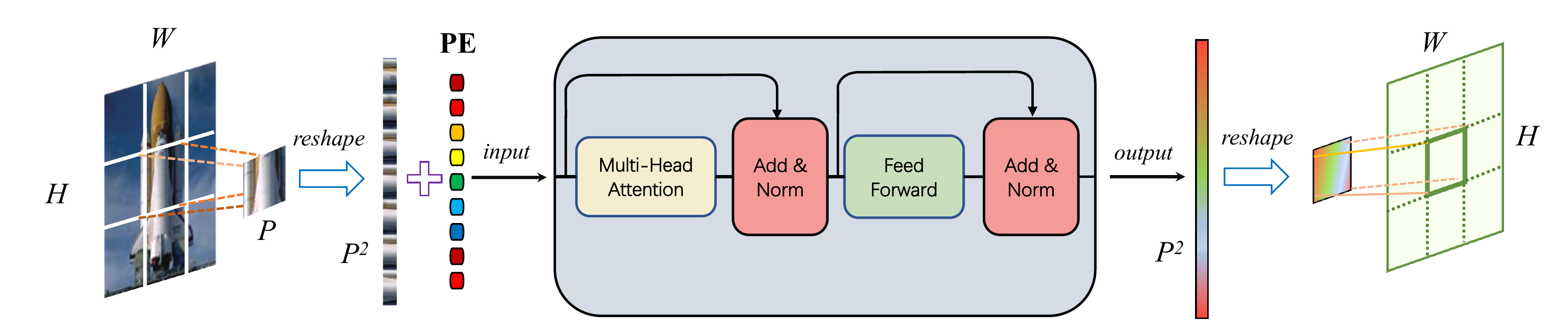}
\end{center}
\caption{A Patch-wise Standard Transformer Encoder (STE) in ConTNet. $\textbf{PE}$ denotes positional encoding. $H$ and $W$ is the height and width of the input and output image separately. $P$ is the size of patch. $P^2$ is the length of the input and output sequence of STE.}
\label{fig: STE in ConTNet}
\end{figure*}







Convolutional networks (ConvNets) are widely used in computer vision and become a dominating method in almost all intelligent vision systems~\cite{krizhevsky2012imagenet, long2015fully, toshev2014human, girshick2015region}. Since most landmark ConvNets mainly use 3x3 convolutions, the receptive field is limited within a local neighbourhood to capture local representation. However, the large receptive filed is of great importance to construct a contextual visual understanding especially in some downstream tasks such as object detection and semantic segmentation. In order to enlarge the receptive field of the ConvNet, stacking multiple convolutional layers (conv layer) seems to be the consensus and induce the flourishing of convolutional backbones useful for various downstream tasks~\cite{simonyan2014very, he2016identity, he2016deep, huang2017densely}.

In Natural Language Processing (NLP), the very major issue is how to model long-range dependencies in long sequences. Self-attention mechanism represented by transformer has become the foremost method and achieved state-of-the-art results on many NLP tasks. The success of transformer~\cite{vaswani2017attention} strongly motivates researchers to utilize transformer in CV tasks~\cite{carion2020end, dosovitskiy2020image, touvron2020training}. However, these vision transformer are highly sensitive to training settings such as learning rate, number of training epochs, optimizer, data augmentation, etc.

We mainly concentrate on the following challenges in CV: (1) ConvNets is deficient in large receptive fields due to the locality of convolution, leading to a performance degradation on downstream tasks. (2) The transformer-based vision model requires special training settings or hundreds of millions of images as the pretraining dataset, which is a practical constraint hampering the widespread adoption.

To overcome these challenges, we propose a novel \textbf{Con}volution-\textbf{T}ransformer \textbf{Net}work (ConTNet) for CV tasks. ConTNet is implemented by stacking multiple ConT blocks as shown in Figure ~\ref{fig: ConT block}. The ConT block treated the standard transformer encoder (STE) as an independent component the same as a conv layer. Specifically, as shown in Figure~\ref{fig: STE in ConTNet} a feature-map is split into several patches of the same size and each patch flattened to a (super) pixel sequence is next sent to STE. Finally the patch embeddings are reshaped back to feature-maps and they are fed into the next conv layer or STE. 

In the proposed ConTNet, the STE plays a leading role in capturing more contextual features, while conv layers efficiently extract local visual information. Besides, we also find that embedding STE into ConvNet architectures can make the network more robust. Or in other words, ConTNet can be trained easily just like the most popular ResNet~\cite{he2016deep}.

We demonstrate that ConTNet is superior to DeiT (transformer-based network) and ResNet (convolution-based network) on ImageNet classification. In accordance with our empirical results, ConTNet can be optimized straightforward like normal ConvNets, and therefore do not requires as many tricks as DeiT~\cite{touvron2020training} or the tremendous amount of pretraining data used by ViT~\cite{dosovitskiy2020image}. Another interesting finding is that ConTNet gains more performance improvement from strong data augmentation and other training tricks than that of ResNet~\cite{he2016deep}, which can be attributed to the overfitting risk of transformer architecture. Some key results are listed below: Our ConT-M achieves a $1.6\%$ top-1 accuracy over ResNet50 on ImageNet~\cite{deng2009imagenet} dataset with almost $25\%$ relative save of computational cost. Likewise, we demonstrate that ConTNet significantly improves dense prediction tasks, especially object detection and segmentation, against the most popular backbone ResNet~\cite{he2016deep}. Our ConT-M yields around $3\%$ improvement compared to ResNet-50~\cite{he2016deep} based on FCOS~\cite{tian2019fcos} and Mask-RCNN~\cite{He_2017_ICCV}.

In a nutshell, this work’s contributions are threefold.
\begin{itemize}
  \item [1)] 
  To our knowledge, our proposed ConTNet is the first exploration to build a neural network with both of the standard transformer encoder (STE) and spatial convolution.
  \item [2)]
  In contrast to the recently trendy visual transformer, ConTNet is much easier to optimize. In addition to robustness, ConTNet performs excellently on image recognition.
  \item [3)]
   The empirical results present that a good transfer learning performance is promised. These results suggest that ConTNet provides a new conv-transformer-based pattern to enlarge model's receptive field. 
\end{itemize}

\section{Related Work}

\subsection{ConvNets}

\begin{figure*}[t]
\begin{center}
   \includegraphics[width=1.0\textwidth]{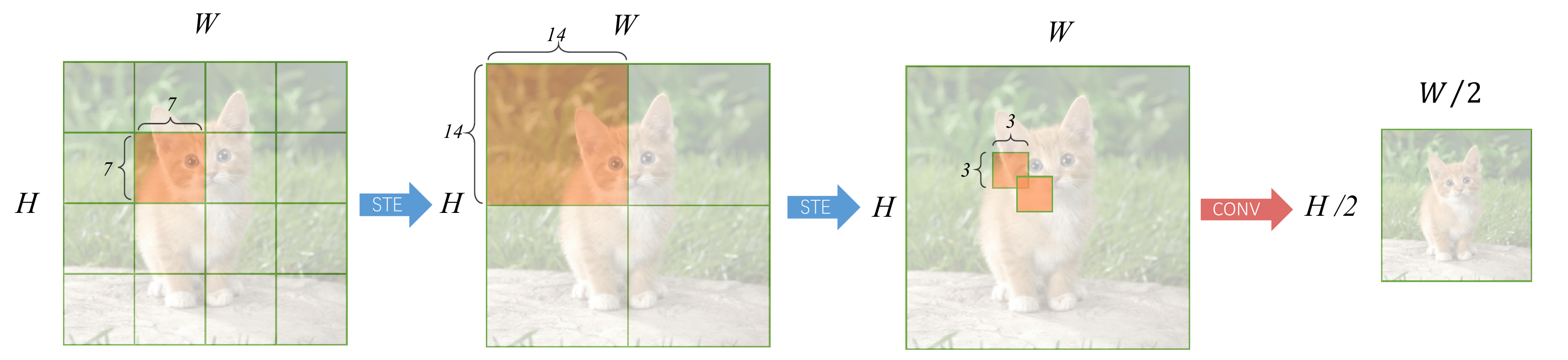}
\end{center}
   \caption{Information flow in a ConT block. For the area covered by orange shadow, the blue arrow indicates that output value of each spatial position is computed from the entire covered area through a patch-wise STE. The red arrow is a conv layer with a kernel size of 3 and a stride of 2, computing only the output value of center position from the entire covered area.}
\label{fig: ConT block true}
\end{figure*}

Deep Convolutional Neural Networks (ConvNets) enables the computer vision (CV) model to perform at a higher level. This decade, we have witnessed that many novel ideas make ConvNets unprecedented powerful, generalizable, and computationally economical~\cite{he2016deep,  huang2017densely, wang2017residual, howard2017mobilenets, kornblith2019better, krizhevsky2012imagenet, chollet2017xception, he2019bag, simonyan2014very, ioffe2015batch}. A succession of researchers improves the performance by deepening the network and adopt the multi-branch architecture that lays the foundation for modern network design~\cite{he2016deep,  huang2017densely, szegedy2016rethinking, xie2017aggregated, he2016identity, zagoruyko2016wide, szegedy2017inception}. An alternative line of practitioners turns to reformulate convolution or basic block as versatile image features~\cite{dai2017deformable, zhu2019deformable, hu2018squeeze, wang2018non, gao2019res2net}. One of these functional attempts is to incorporate self-attention mechanism into ConvNets, which promotes global features utilization of ConvNets~\cite{wang2018non, hu2018gather, cao2019gcnet, park2018bam, woo2018cbam, hu2018squeeze}. SENet~\cite{hu2018squeeze} models the interdependencies between channels according to the global context of feature-map. GENet~\cite{hu2018gather} is a generalization of SENet, still first gathering global context as a signal to recalibrate the feature channel. Like CBAM~\cite{woo2018cbam} and BAM~\cite{park2018bam}, multiple papers rescale both of different spatial positions and channels by aggregating all contextual information. Non-Local Net~\cite{wang2018non} implements a pair-wise pixel interaction to augment the long-range dependencies across all temporal frames and spatial positions, and Non-local block can introduce the self-attention mechanism. However, the usage of self-attention by Non-Local Net has weaknesses, for instance, the Non-Local Block is hard-weight and insensitive to position of pixels. Another series of works seek to adopt self-attention mechanism along the entire network or even replacing all spatial convolution with self-attention to construct an efficient fully attentional network~\cite{ramachandran2019stand, hu2019local, bello2019attention, zhao2020exploring}. These efforts also aim to enhance the long-range dependencies of ConvNets.

Nevertheless, many of these attentional networks are designed elaborately and have not shown strong applicability for downstream tasks. Based on these observations, we innovatively develop ConTNet combining standard transformer encoder (STE) together with convolution layer (conv layer). The network design follows the principle of STE for global features and conv layer for local features. Such an architecture provides a reasonable formula for elevating the network's ability to model long-range dependencies.

\subsection{Transformer}

Transformer~\cite{vaswani2017attention} is an encoder-decoder neural network for sequence-to-sequence tasks, which has achieved many state-of-the-art results and further revolutionized NLP with the success of BERT~\cite{devlin2018bert}. The recently trendy visual transformer has shown that an end-to-end standard transformer can implement image classification and other vision tasks~\cite{carion2020end, liu2020convtransformer, jiang2021transgan, zheng2020rethinking, zhu2020deformable}. ViT~\cite{dosovitskiy2020image} cuts the images into some non-overlapping patches and encodes the patches set as a token sequence, whose head is attached to a learnable classification token. The performance of ViT depends on large-scale pretrain datasets like ImageNet-21k dataset or JFT-300M dataset, which trumps the convolutional inductive bias. DeiT~\cite{touvron2020training} explores distillation to extend ViT to a data-efficient vision transformer straightly trained on ImageNet, but the training course of DeiT is complex and unstable. Transformer has also been extended to solving dense prediction problems or low-level tasks. For example, DETR~\cite{carion2020end} is the first work to using transformer for object detection. DETR uses ConvNets to extract features and uses transformer to model the object detection as an end-to-end dictionary lookup problem. However, DETR is still very sensitive to the hyper-parameters setting and requires a longer training period. 

In our practice, embedding STEs into ConvNet, which seems to alternately employ transformer and convolution, can make transformer architecture as robust as convolution (see Figure~\ref{fig: STE in ConTNet}). We combine transformer and convolution as a ConT block, and ConTNet is formed by stacking ConT blocks as shown in Figure~\ref{fig: ConT block}. ConTNet can be trained from scratch on ImageNet dataset and generalized to dense prediction tasks without unusual settings as expected.

\begin{table*}[]
\begin{center}
\begin{tabular}{@{}cccccc@{}}
\toprule
Stage    & Input size & ConT-Ti                & ConT-S               & ConT-M               & ConT-B               
\\
\midrule
Stage 0 & 224 $\times$ 224       & \multicolumn{4}{c}{ 7$\times$7,\ 64,\ \text{stride}=2,\  \text{padding}=3  }                                                      
\\

\midrule
& & & & &
\\
Stage 1 & $56\times56$     &  $\left[ \begin{tabular}[c]{@{}l@{}} $D=48$, \\ $D_{ffn} =192$, \\ $H=1$\end{tabular} \right]  \times 1$    & $\left[ \begin{tabular}[c]{@{}l@{}} $D=64$, \\ $D_{ffn}=256$, \\ $H=1$ \end{tabular} \right] \times 1$   & $\left[ \begin{tabular}[c]{@{}l@{}}$ D=64$, \\ $D_{ffn}=256$, \\ $H=1$\end{tabular} \right] \times 2$   & $\left[ \begin{tabular}[c]{@{}l@{}}$D=64$, \\ $D_{ffn}=256$, \\ $H=1$ \end{tabular} \right] \times 3$   
\\
& & & & &
\\

\midrule

& & & & &
\\
Stage 2 & $ 28\times28    $   & $\left[ \begin{tabular}[c]{@{}l@{}} $D=96$, \\ $D_{ffn}=384$, \\ $H=2 $\end{tabular} \right] \times 1$     & $\left[ \begin{tabular}[c]{@{}l@{}}$D=128$, \\$ D_{ffn}=512 $, \\ $ H=2 $\end{tabular} \right] \times 1$  & $\left[ \begin{tabular}[c]{@{}l@{}}$D=128$, \\ $D_{ffn}=512$, \\ $H=2$ \end{tabular} \right] \times 2$  & $\left[ \begin{tabular}[c]{@{}l@{}}$D=128$, \\ $D_{ffn}=512$, \\ $H=2$\end{tabular} \right] \times 4$  
\\
& & & & &
\\

\midrule

& & & & &
\\
Stage 3 & $ 14\times14 $        & $\left[ \begin{tabular}[c]{@{}l@{}}$ D=192 $, \\ $ D_{ffn}=768 $, \\ $ H=4 $\end{tabular} \right] \times 1$    & $\left[ \begin{tabular}[c]{@{}l@{}} $ D=256$ , \\ $ D_{ffn}=1024 $, \\ $ H=4 $\end{tabular} \right] \times 1$ & $\left[ \begin{tabular}[c]{@{}l@{}} $ D=256 $, \\ $ D_{ffn}=1024 $ , \\$ H=4 $\end{tabular} \right] \times 2$ & $\left[ \begin{tabular}[c]{@{}l@{}} $ D=256 $, \\ $ D_{ffn}=1024 $, \\ $ H=4$ \end{tabular} \right] \times 6$ 
\\
& & & & &
\\

\midrule
& & & & &
\\
Stage 4 & $ 7\times7 $         & $\left[ \begin{tabular}[c]{@{}l@{}} $D=384$ , \\ $D_{ffn}=768$, \\ $H=8$\end{tabular} \right] \times 1$    & $\left[ \begin{tabular}[c]{@{}l@{}}$D=512$ , \\ $D_{ffn}=1024$, \\ $ H=8 $\end{tabular} \right] \times 1$ & $\left[ \begin{tabular}[c]{@{}l@{}} $D=512$, \\$ D_{ffn}=1024 $, \\ $ H=8 $\end{tabular} \right] \times 2$ & $\left[ \begin{tabular}[c]{@{}l@{}} $ D=512$ , \\ $ D_{ffn}=1024 $, \\ $ H=8 $\end{tabular} \right] \times 3 $
\\
& & & & &
\\
\midrule
& $ 7\times7 $ & \multicolumn{4}{c}{global average pooling, 1000-d fc, softmax}  
\\
\bottomrule
\end{tabular}
\end{center}
\caption{Detailed settings of ConTNet series. Inside the brackets, we list the hyper-parameter of each ConTBlock. $D$ is the embedding dimension of MHSA, $D_{ffn}$ is the dimension of FFN, and $H$ is the head number of MHSA. Outside the brackets, the number of stacked blocks on the stage is presented. In each stage, the last conv layer performs downsampling and increases dimension.
}
\label{ConTNet architectural table}
\end{table*}
\section{ConTNet}

In this section, we describe the proposed \textbf{Con}volution-\textbf{T}ransformer \textbf{Net}work (ConTNet) in details.

\subsection{Network architecture}
We propose the novel convolution-transformer-based network ConTNet. The structure of ConTNet is shown in Figure \ref{fig: ConT block}. ConTNet is composed of standard transformer encoders (STEs) and spatial convolutions that are stacked seemingly alternately. More precisely, our very first step towards ConTNet is setting up a relatively shallow ConvNet, which has four stages processing feature-maps with different sizes and channels. This ConvNet is then extended to ConTNet through inserting STE between two neighboring convolutional (conv) layers. Such an extension is made to capture global features supplementary to local representations learned by conv layers.

To systematically embedding STEs into ConvNet, we design a block make STE fully integrated with conv layers by grouping them in pairs. A ConT block cascades a pair of STEs and a conv layer as shown in Figure~\ref{fig: ConT block}. Each ConT block comprises two STEs and one conv layer with a kernel size of $3\times 3$. In our implementation, the spatial size of split patches is set to $7$ and $14$ sequentially. Inspired by ResNet, we construct a shortcut connection to implement the residual learning $y = f(x)+x$, which is widely utilized to improve the performance of ConvNets. ConTNet is still a 4-stage-style hierarchical network because we desire it to be suitable for downstream tasks, especially object detection and segmentation. In each stage, the last conv layer of the stage has a stride of 2 to conduct downsampling and increasing channels. When trained on the ImageNet dataset, the size of each stage's feature-map is $ [ 56, 28, 14, 7 ] $, and the channel of each stage is determined by scaling the popular setting $[64, 128, 256, 512]$ used by most ConvNets~\cite{he2016deep}. The head number of multi-head self attention of STE in each stage is set to $[1,2,4,8]$ to keep the channel of single-head attention $64$. Note that all conv layers in ConTNet have a kernel of $3 \times 3$ except for the top and the bottom conv layer. A conv layer with a kernel of $7 \times 7$ and a MaxPooling are placed at the top of the network, which follows the practice of ResNet. And the last conv layer of the last stage has a kernel of $1 \times 1$ to save parameters.

To produce architectural variants of ConTNet, we scale the depth and width of ConTNet to reach different computational budgets. Table \ref{ConTNet architectural table} presents four architectures: ConT-Ti (Tiny), ConT-S (Small), ConT-M (Medium), ConT-B (Big). From ConT-Ti to ConT-B, the parameters and FLOPs are gradually increased, and the depth or width grows progressively.

\subsection{Patch-wise Standard Transformer Encoder} \label{subsection32}
In this subsection, we revisit the Standard Transformer Encoder (STE) and present a procedure for the execution of its capturing long-range dependency in ConTNet. 

In the raw application of standard Transformer~\cite{vaswani2017attention}, a sequence of words is received as input and finally translated into a new sentence. When applied to Computer Vision (CV) tasks, Transformer encoder has to take a 2D image as input. On the principle of retaining authentic STE, the input 2D image $ \mathbf{x_{\rm{2d}}} \in \mathbb{R} ^ \mathit{H \times W \times C}$ is flattened into a sequence of pixels denoted by $ \mathbf{x_{\rm{1d}}} \in \mathbb{R} ^ \mathit{(H \times W) \times C}$, where $H$ and $W$ are height and width of the input image separately, and $C$ is the image channel. Towards a simple but powerful utilization of STE, we develop a method that roughly similar to convolutional filter (see Figure~\ref{fig: STE in ConTNet}). Given a 2D-image $ \mathbf{x_{\rm{2d}}} \in \mathbb{R} ^ \mathit{H \times W \times C_{in}}$, a convolutional filter aggregates spatial information across a local neighbourhood. For example, a kernel of shape $ 3\times 3\times C_{in}$ slides over every position of image to do a inner product of local window and the kernel weights. Assuming that the input $ \mathbf{x_{\rm{2d}}}$ and the output $ \mathbf{y_{\rm{2d}}} \in \mathbb{R} ^ \mathit{H \times W \times C_{out}}$ has the same spatial size, for a spatial position $x_{ij; i \in [0, H), j\in[0, W)}$, the corresponding output value $y_{ij; i \in [0, H), j\in[0, W)}$ is calculated as follow: 

\begin{equation}
y_{ij} = \rm{Conv}\left( \mathit{x_{ij}} \right)  \label {XX},
\end{equation}
\begin{equation}
 \rm{Conv}\left( \mathit{x_{ij}} \right)={\sum_{\mathit{a,b=-k}}^{\mathit{k}}\mathit{W_{a,b}}\mathit{x_{i+a, j+b}}},
\end{equation}
where $W \in \mathbb{R} ^ \mathit{\left(2k+1\right) \times \left(2k+1\right) \times C_{in} \times C_{out}}$ is the weight of convlutional filter, and $a,b \in \left[-k, k \right]$. 
With above analysis, a pixel-to-pixel mapping is supported explicitly in convolutional filter. By contrast, the STE learns a mapping from sequence to sequence, as shown in Figure~\ref{fig: STE in ConTNet}. Therefore, a potential way to exploit the STE in a conv layer fashion is considering a sequence as a pixel. Prior to performing on an image, STE first splits it into several patches of the same size $P \times P $, and treats each patch as a sequence of pixels. The split images can be denoted by a new tensor $ \mathbf{x_{\rm{2d}}^{\rm{p}}} \in \mathbb{R} ^ \mathit{H_p \times W_p \times P^2 \times C}$, where $H_p$ is set to $H \slash p$ and $W_p$ is set to $W \slash p$. Instead of that each spatial position of $ \mathbf{x_{\rm{2d}}} $ is a pixel, each spatial position of $ \mathbf{x^{\rm{p}}_{\rm{2d}}} $ is a sequence $x^{p}_{mn; m \in [0, H_p), n\in [0, W_p)} \in \mathbb{R} ^ \mathit{P^2 \times C}$. Assuming that the input $ \mathbf{x^{\rm{p}}_{\rm{2d}}} $  has the same spatial size and channel as the output $ \mathbf{y^{\rm{p}}_{\rm{2d}}} $, for a spatial position $(m, n)$, the output value $y^{p}_{mn; m \in [0, H_p), n\in [0, W_p)} \in \mathbb{R} ^ \mathit{P^2 \times C}$ is calculated as follow:
\begin{equation}
y_{mn}^{p} = \rm{STE}\left( \mathit{x_{mn}^{p}} \right).  \label {XX}
\end{equation}
The operation of STE is enumerated exactly according to ~\cite{vaswani2017attention}, which can be formulated:
\begin{equation}
    \rm{STE}\left( \mathit{x_{mn}^p} \right) = \rm{FFN} \left(\rm{MHSA}\left(\mathit{x_{mn}^{p}}+\mathbf{PE} \right) \right),
\end{equation}
where $\rm{MHSA} \left( \space \right)$ is a Multi-Head Self Attention (MHSA) mechanism and $\rm{FFN} \left( \space \right)$ is a 2-layer Feed Forward Network (FFN), and $\mathbf{PE} \in \mathbb{R} ^ \mathit{P^2 \times C}$ is positional encoding. 

Finally, we descirbe a principled execution of MHSA and FFN. Take a sequence $ \mathbf{x_{\rm{1d}}} \in \mathbb{R} ^ \mathit{N \times C}$ as input, a single-head attention value is computed using:
\begin{equation}
A=\rm{softmax} \left( \frac{ {(\mathbf{W_{\rm{q}}}\mathbf{x_{\rm{1d}}})}{(\mathbf{W_{\rm{k}}}\mathbf{x_{\rm{1d}}})^{\rm{T}}}}{ \sqrt{\mathit{D_{\rm{h}}}}}\right) (\mathbf{W_{\rm{v}}}\mathbf{x_{\rm{1d}}}),
\end{equation}
where $\mathbf{W_{\rm{q}}}$, $\mathbf{W_{\rm{k}}} $ and $\mathbf{W_{\rm{v}}}  \in \mathbb{R} ^ \mathit{C\times D_h}$ are the learned linear transformations, and $D_h$ is typically set to $D \slash h$. $D$ is the embedding dimension and $h$ is the number of head. Multi-head attention value is obtained by projecting concatenated single attention values:
\begin{equation}
A_{mh}=\left[\mathit{A_1};\mathit{A_2};...;\mathit{A_h}\right] {\mathbf{W_{\rm{mhsa}}}},
\end{equation}

\begin{equation}
    \rm{MHSA}\left(\mathbf{x_{\rm{1d}}}\right) = \rm{LN}\left(\mathit{A_{mh}} + \mathbf{x_{\rm{1d}}} \right),
\end{equation}
where $\mathbf{W_{\rm{mhsa}}}  \in \mathbb{R} ^ \mathit{D\times C}$ is the learned weights that aggregates multiple attention values, and $\rm LN \left( \right)$ is Layernorm~\cite{ba2016layer}, applied after a residual connection. The output of MHSA is fed into FFN:
\begin{equation}
    \rm {FFN} \left(\mathbf{x_{\rm{1d}}} \right) = \rm{LN}\left(\mathbf{W_{\rm{2}}} \mathbf{W_{\rm{1}}} \mathbf{x_{\rm{1d}}}  +\mathbf{x_{\rm{1d}}}\right),
\end{equation}
where $\mathbf{W_{\rm{1}}} \in \mathbb{R} ^ \mathit{C\times D_{ffn}}$ and $\mathbf{W_{\rm{2}}} \in \mathbb{R} ^ \mathit{ D_{ffn} \times C}$ are both learned linear transformations.


\subsection{Analysis and More Details}
We have revealed that $ \left(1 \right)$ the STE is adopted on each sequence $x^{p}_{mn; m \in [0, H_p), n\in [0, W_p)} \in \mathbb{R} ^ \mathit{P^2 \times C}$ which is flattened from the uniformly split patch by sliding the window on the separate input image $ \mathbf{x_{\rm{2d}}^{\rm{p}}} \in \mathbb{R} ^ \mathit{H_p \times W_p \times P^2 \times C}$ and $ \left(2 \right)$ ConTNet captures both of global and local features by filtering feature-maps with conv layers and STEs alternately. More details about these two implementations are reported in this subsection.

\noindent
\medskip\\
\textbf{Using STE like a kernel:} In ConTNet, an STE slides over the image and translates each split patch into a new patch, which performs like a filter with kernel size and stride both equal to the patch size. We find such an kernel-like operation has two favourable properties. Concretely speaking, the patch-wise STE is weight-shared, which has translation equivariance and computational efficiency. Instead of a pixel-wise translation equivariance, a relatively rough patch-wise translation equivariance is obtained through our operation. For each patch, some work endeavors to model a pixel-wise translation equivariance when performing a self-attention mechanism on a 2D-shape image. To alleviate this issue, we choose a simple but effective method: A 
parameter-shared learned PE is added to each split 
patch, recording each pixel's coordinate in a patch and avoid permutation equivariance problem~\cite{vaswani2017attention}. In contrast with a conv layer with a kernel of $3 \times 3$ that has $\mathit{9}{C^2}$ parameters and a computational complexity of $\mathit{9}{C^2HW}$ (assuming that the input and the output are of the same channel) , an STE has ${2}{D_{mhsa}}{D_{ffn}} + {4}{D_{mhsa}^2}+{{P^2}{D_{mhsa}}}$ parameters and a computational complexity of ${2}{D_{mhsa}}{D_{ffn}{HW}} + {4}{D_{mhsa}^2{HW}}+{\left(HW\right)/{P^2}}$. When the dimension of MHSA ${D_{mhsa}}$ and the input's channel ${C_{in}}$ are equal and the dimension of FFN ${D_{ffn}}$ is equal to four times ${D_{mhsa}}$ (practice of~\cite{vaswani2017attention}), the increase of parameters is ${3}{C^2}+{{P^2}{C}}$ and the increase of computational compexity is ${3}{C^2}{HW}+{\left(HW\right)/{P^2}}$.

\noindent
\medskip\\
\textbf{Alternately capturing features with STE and conv layer:} 
 ConvNets are built upon multiple conv layers with an advantage of locality as well as a lack of global features. To take full advantage of Transformer to enlarge receptive fields, we build ConT block, the basic block of ConTNet, containing two STEs and one conv layer. In each ConT block, STEs first capture more global features, and then a conv layer with a small kernel captures more local features. Hence, ConTNet interweaves the features alternately captured by STEs and conv layers via stacking ConT blocks.
Owing to the patch-wise operation on the STE, one pixel has interactions with all pixels in its patch, and we hypothesize the size of the split patch can be adjusted flexibly to model kernels of different receptive fields. We introduce a dynamic setting of patch size rather than a fixed version along the entire network. The default patch size of the first and the second STE is set to 7 and 14 separately, and the kernel size of conv layer is set to 3 (see Figure~\ref{fig: ConT block true}). An identity shortcut is connected between the input and output of the ConT block by doing an element-wise addition. To conduct downsampling and dimension changing, the conv layer of the last ConT block in each stage has a stride of 2 and increases dimension. In this special case, a projection shortcut replaces the identity shortcut, using a $1 \times 1$ conv layer with a stride of 2 to match the increased dimension and smaller size for element-wise addition.


\section{Experiments}

In this section, we conduct extensive experiments on image classification and downstream tasks to evaluate the effectiveness of our ConTNet.

\subsection{Image Classification}
We compare ConTNet to advanced ConvNets (ResNet~\cite{he2016deep}) and transformer-based backbone (DeiT~\cite{touvron2020training}) separately to assess the image classification performance on ImageNet dataset~\cite{deng2009imagenet}. ImageNet dataset contains a train set of 1.28 million images and a validation set of 50000 images. We train all models on the train set and test performance on the validation set. ResNet is the chosen competitor against ConTNet since it is the most popular ConvNet in CV applications. On the other hand, we consider DeiT as representative of the visual transformer, which can be directly trained from scratch on ImageNet dataset.


\begin{table}[]
\begin{center}
\setlength{\tabcolsep}{4mm}{
\begin{tabular}{@{}lccc@{}}
\toprule
Network & FLOPs(G) & \#Param(M) & Top-1(\%)      \\
\midrule
Res-18  & 1.8      & 11.7    & 71.5          \\
ConT-S  & 1.5      & 10.1    & \textbf{74.9} \\
Res-50  & 4.0      & 25.6    & 77.1          \\
ConT-M  & 3.1      & 19.2    & \textbf{77.6} \\
Res-101 & 7.6      & 44.5    & \textbf{78.2} \\
ConT-B  & 6.4      & 39.6    & 77.9         \\
\bottomrule
\end{tabular}
}
\end{center}
\caption{Top1 accuracy $\left(\% \right)$ on ImageNet. ResNet and ConTNet are trained with the same setting.}
\label{imagenet cont vs. res}
\end{table}

\begin{table}[]
\begin{center}
\setlength{\tabcolsep}{4mm}{
\begin{tabular}{@{}lccc@{}}
\toprule
Network & FLOPs(G) & \#Param(M) & Top-1(\%)  \\
\midrule
Res-18\textsuperscript{*}                    & 1.8                          & 11.7       & 73.2                         \\
ConT-S\textsuperscript{*}                    & 1.5                          & 10.1       & \textbf{76.5}                         \\
Res-50*                     & 4.0                          & 25.6       & 78.6                         \\
ConT-M\textsuperscript{*}                     & 3.1                          & 19.2       & \textbf{80.2}                         \\
Res-101\textsuperscript{*}                    & 7.6                          & 44.5       & 80.0                         \\
ConT-B\textsuperscript{*}                & 6.4                          & 39.6       & \textbf{81.8}     \\
\bottomrule
\end{tabular}}
\end{center}
\caption{Top1 accuracy $\left(\%\right)$ on ImageNet. $*$ indicates that the model is trained with strong data augmentation containing mixup and auto-augmentation. AdamW is used to optimize ConTNet here instead of SGD. The fairness of comparsion is enabled by the fact that SGD is fairly good for ResNet.}
\label{ImageNet aug}
\end{table}

\begin{table}[]
\begin{center}
\setlength{\tabcolsep}{4mm}{
\begin{tabular}{@{}lccc@{}}
\toprule
Network  & FLOPs(G) & \#Param(M) & Top-1(\%) \\
\midrule
DeiT-Ti  & 1.3         & 5.7          & 72.2     \\
ConT-Ti\textsuperscript{*} & 0.8      & 5.8        & \textbf{74.9}     \\
DeiT-S   & 4.6     & 22.1         & 79.8     \\
ConT-M\textsuperscript{*}  & 3.1      & 19.2       & \textbf{80.2}     \\
DeiT-B   & 17.6     & 86.6         & \textbf{81.8}     \\
ConT-B\textsuperscript{*}  & 6.4      & 39.6       & \textbf{81.8}    \\
\bottomrule
\end{tabular}}
\end{center}
\caption{Top1 accuracy $\left(\%\right)$ on ImageNet. Results of DeiT is from the original paper. $*$ indicates that model is trained with mixup and auto-augmentation with AdamW optimizer.}
\label{Imagenet Cont vs. DeiT}
\end{table}

\begin{table*}[]
\begin{center}
\setlength{\tabcolsep}{1.2mm}
\begin{tabular}{@{}llccllllll@{}}
\toprule
Method                     & Backbone & Param(M) & FLOPs(G) & AP & $AP_{50}$ & $AP_{75}$ & $AP_s$ & $AP_m$ & $AP_l$ \\
\midrule
{RetinaNet} & Res-50   & 32.0  & 235.6  &  36.5       & 55.4       & 39.1       & 20.4       & 40.3       & 48.1       \\
                     & ConT-m   &  27.0 & 217.2  &  37.9(\color[RGB]{0,100,0}{\textbf{+1.4}}) & 58.1(\color[RGB]{0,100,0}{\textbf{+2.7}}) & 40.2(\color[RGB]{0,100,0}{\textbf{+1.1}}) & 23.0(\color[RGB]{0,100,0}{\textbf{+2.6}}) & 40.6(\color[RGB]{0,100,0}{\textbf{+0.3}}) & 50.4(\color[RGB]{0,100,0}{\textbf{+2.3}}) \\
{FCOS}     & Res-50    &  32.2 & 242.9  &  38.7       & 57.4       & 41.8       & 22.9       & 42.5       & 50.1       \\
                    & ConT-m    & 27.2  &  228.4 &  40.8(\color[RGB]{0,100,0}{\textbf{+2.1}}) & 60.5(\color[RGB]{0,100,0}{\textbf{+3.1}}) & 44.7(\color[RGB]{0,100,0}{\textbf{+2.9}}) & 25.1(\color[RGB]{0,100,0}{\textbf{+2.2}}) & 44.6(\color[RGB]{0,100,0}{\textbf{+2.1}}) & 53.0(\color[RGB]{0,100,0}{\textbf{+2.9}}) \\
{Faster RCNN}     & Res-50    & 41.5  &  241.0 &  37.4       & 58.1       & 40.4        & 21.2       & 41.0       & 48.1       \\
                           & ConT-m    & 36.6  & 225.6  &  40.0(\color[RGB]{0,100,0}{\textbf{+2.6}}) & 62.4(\color[RGB]{0,100,0}{\textbf{+4.3}}) & 43.0(\color[RGB]{0,100,0}{\textbf{+2.6}})  & 25.4(\color[RGB]{0,100,0}{\textbf{+4.2}}) & 43.0(\color[RGB]{0,100,0}{\textbf{+1.9}}) & 52.0(\color[RGB]{0,100,0}{\textbf{+3.9}}) \\
\bottomrule
\end{tabular}

\end{center}
\caption{Object detection mAP $\left( \% \right)$ on the COCO validation set. We evaluate the performance of detection model by replacing backbone ResNet50 in ReTinaNet, FCOS and Faster-RCNN with ConTNet.}
\label{object detection results}
\end{table*}

Towards a fair competition, we trained ConTNet and ResNet with identical training settings. We use SGD~\cite{sra2012optimization} to train both of ConTNet and ResNet with a batch-size of 512 for 200 epochs on 8 Nvidia V100 GPUs. Cosine decay schedule is used to adjust learning rate. The initial learning rate is set to 0.2 for ResNet and conv layers in ConTNet, and the initial learning rate of STEs in ConTNet is uniquely set to 0.005. We use a weight decay of 0.00005 and a momentum of 0.9. We follow a simple data-augmentation in ResNet and perform regularization by using label smooth with $\epsilon$ of 0.1. Table \ref{imagenet cont vs. res} shows the results of comparison. Specifically, ConT-S ourperforms ResNet-18 (by $3.4\%$), and ConT-M performs sligtly better than ResNet-50 (by $0.5\%$), while saving nearly $20\%$ parameters as well as $25\%$ computational costs. However, ConT-B achieves a slightly lower accuracy than ResNet-101 (by $-0.3\%$), as the computational complexity is lower. We hypothesize that STE in ConTNet has a high risk of overfitting. To eliminate the risk, we train ConTNet and ResNet both with strong data augmentations including auto-augmentation~\cite{cubuk2018autoaugment} and mixup~\cite{zhang2017mixup}. Additionally, we use AdamW~\cite{loshchilov2018fixing} optimizer in stead of SGD to train ConTNet with an initial learning rate of 0.0005 and a weight decay of 0.05. Table \ref{ImageNet aug} shows the results with strong augmentations. In the case of using the same strong data augmentation methods, ConTNet earns more benefits than ResNet, and all ConTNets outperform ResNets on ImageNet dataset across different computational budgets. For example, ConT-B achieves $81.8\%$, which is $1.8\%$ better than Res-101.

\begin{table*}[]
\begin{center}
\setlength{\tabcolsep}{2.5mm}{
\begin{tabular}{@{}lllllllll@{}}
\toprule
Backbone & $AP^{bb}$       & $AP^{bb}_{s}$      & $AP^{bb}_{m}$      & $AP^{bb}_{l}$      & $AP^{mk}$       & $AP^{mk}_s$      & $AP^{mk}_m$      & $AP^{mk}_l$      \\
\midrule
Res-50   & 38.2       & 21.9       & 40.9       & 49.5       & 34.7       & 18.3       & 37.4       & 47.2       \\
ConT-M   & 40.5(\color[RGB]{0,100,0}{\textbf{+2.3}}) & 25.1(\color[RGB]{0,100,0}{\textbf{+3.2}}) & 44.4(\color[RGB]{0,100,0}{\textbf{+3.5}}) & 52.7(\color[RGB]{0,100,0}{\textbf{+3.2}}) & 38.1(\color[RGB]{0,100,0}{\textbf{+3.4}}) & 20.9(\color[RGB]{0,100,0}{\textbf{+2.6}}) & 41.0(\color[RGB]{0,100,0}{\textbf{+3.6}}) & 50.3(\color[RGB]{0,100,0}{\textbf{+3.1}})
\\
\bottomrule
\end{tabular}}
\end{center}
\caption{Instance segmentation mAP $\left( \% \right)$ on the COCO validation set. We replace the ResNet50 of Mask-RCNN with ConT-M.}
\label{instance seg results}
\end{table*}

The results of DeiT presented in Table \ref{Imagenet Cont vs. DeiT} is sourced from the original paper, which is trained with a pipeline of strong data augmentations and many tricks with AdamW optimizer~\cite{touvron2020training}. With the same optimizer, fewer data augmentations and other tricks, ConTNet presents a better performance against DeiT. For example, ConT-B achieves $81.8\%$, the same as DeiT-B, while our FLOPs is $60\%$ fewer (6.4 G $vs.$ 17.6 G). 

One can also find that the performance of DeiT is extremely sensitive to the data augmentations and training tricks~\cite{touvron2020training}. For instance, without either stochastic depth or random erasing, the convergence of DeiT is seemingly disabled. If traind with a SGD optimizer instead of AdamW, DeiT achieves a much lower accuracy by 7.3. By contrast, our experiments have shown that ConTNet can be optimized stably even in the same setting as ResNet and only a few data augmentation tricks incur a better accuracy exceeding DeiT, which verifies that ConTNet has an advantage of robustness over DeiT.

\subsection{Downstream tasks}
Downstream tasks are more sensitive to the size of the receptive field, and models capable of capturing long-range dependencies would achieve better results. To demonstrate ConTNet has a stronger ability to capture global information, we make use of ConT-M as the backbone to implement some downstream tasks, including object detection, instance segmentation and semantic segmentation. Our empirical results show that ConT-M obtains a better performance with lower computational complexity in contrast to using resnet50 as backbone.


\noindent
\medskip\\
\textbf{Object Detection:} We examine the object detection task on the COCO2017~\cite{lin2014microsoft} dataset containing a train set of 118k images and a validation set of 5k images. We adopt Faster-RCNN~\cite{ren2015faster} (two-stage), FCOS~\cite{tian2019fcos} (one-stage) and RetinaNet~\cite{lin2017focal} (one-stage) as the detection model with ResNet-50 backbone. Both ConT-M and ResNet-50 are pretrained on ImageNet dataset with the same training setting. Then all models are trained through mmdetection following the same training setting. We modify the input size slightly from $(1333, 800)$ to $(1344, 784)$ and employ marginal padding between stage 2 and 3 to match the patch size of STE. Because the ConTNet conducts downsampling with the last conv layer of each stage, we tweak the stride from 2 to 1 and add Average-Pooling layer for downsampling in order to take full use of FPN~\cite{lin2017feature}. 

Table \ref{object detection results} records the results of detection experiments. The ConTNet-based model outperforms baseline in all object detection methods. Take a close look at these results, $AP_l$ is improved most largely, and $AP_s$ has a higher performance improvement than $A_m$. This observation proves that ConTNet has a relatively large receptive field to capture more global features without 
compromising locality. We conclude that using STE and conv layer alternately can learn more contextual representation than ConvNet while maintaining local features. Most notably, such improvements are attained without any bells and whistles, thus providing an evidence that ConTNet is applicable and stable for the downstream task.
\begin{table}[]
\begin{center}
\setlength{\tabcolsep}{6mm}{
\begin{tabular}{@{}lc@{}}

\toprule
Model      & mIOU(\%)  \\
\midrule
PSP-Res50\textsuperscript{$\star$} & 77.30  \\
PSP-Res50 & 77.12 \\
PSP-ConTM  & 78.28 \\
\bottomrule
\end{tabular}}
\end{center}
\caption{mIOU on cityscapes validation set. ${\star}$ indicates the results of original implementation.}
\label{seg}
\end{table}

\noindent
\medskip\\
\textbf{Instance Segmentation:} We use COCO2017 instance dataset to investigate the performance of ConTNet on instance segmentation. We employ Mask-RCNN~\cite{He_2017_ICCV} with ResNet50 as baseline, following the training settings in object detection experiments. Table \ref{instance seg results} shows that ConTNet improves the $bbox_{map}$ by $2.3\%$ and the $segm_{map}$ by $3.4\%$. The experiments on instance segmentation also support the conclusion proposed in the object detection subsection. When ConTNet serves as the backbone, ConTNet improves the model's performance and captures global features better.

\noindent
\medskip\\
\textbf{Semantic Segmentation:}
We conduct semantic segmentation experiments on the Cityscapes dataset~\cite{cordts2016cityscapes}. We use PSPNet~\cite{zhao2017pyramid} for this implementation. We replace the ResNet50 in PSPNet with ConT-M and follow training and validation protocols in PSPNet except for input size and optimizer. The original input size is$ \left( 713, 713 \right)$ and we use a input size of $ \left(672, 672 \right) $ instead to match split operation of STE. We use AdamW optimizer to train our model on training set and evaluate the performance on validation set. Table \ref{seg} shows that our model achieves a higher mIOU than baseline ($1.16\%$).

\subsection{Ablation study}
\noindent
\textbf{Positional Encoding:}
Positional encoding (PE) is essential to Transformer in NLP tasks. The order of words defines the grammar for a sentence, and the order of pixels determines the semantics of an image. In ConTNet, a standard transformer encoder (STE) infuses a shared learnable 2D position encoding into each split patch. Table~\ref{pe} shows the results of different position encoding schemes. Expectedly, the infusion of PE increases performance of a position-unware ConTNet significantly. Compared to adopting 2D position encoding~\cite{parmar2018image}, the performance degrades slightly using 1D position encoding. We next replace self-attention in STE with relative attention from~\cite{shaw2018self} to implement a relative position encoding, which yields an additional improvement. 

As mentioned above, ConTNet adopts a patch-wise positional infusion by adding a shared PE to each split patch. We are curious aboout the effect of adding PE to entire image before splitting it into patch. Table~\ref{pe} shows that patch-wise position encoding outperforms image-wise position encoding (by $0.9\%$). The result suggests that PE should be used in combination with patch-wise STE more than in isolation. 

\begin{table}[]
\begin{center}
\setlength{\tabcolsep}{6.6mm}{
\begin{tabular}{@{}llc@{}}
\toprule
Position Encoding             & Placement  & Top-1(\%) \\
\midrule
None                          & None       & 78.9      \\ 
1D learnable                  & patch-wise & 80.1      \\ 
2D learnable                  & patch-wise & 80.2     \\ 
2D learnable                  & image-wise & 79.3      \\ 
Relative                      & patch-wise & \textbf{80.4}      \\ 
\bottomrule
\end{tabular}
}
\end{center}
\caption{Results of different position encoding schemes on ImageNet classification. In the case of 2D learnable postion encoding, results of different placements are listed. }
\label{pe}
\end{table}

\begin{table}[]
\begin{center}
\setlength{\tabcolsep}{9mm}{
\begin{tabular}{@{}lc@{}}
\toprule
Patch size                  & Top-1(\%) \\
\midrule
{[} 7, 7{]}, {[} 7, 7{]}, ..., {[} 7, 7{]}, {[} 7, 7{]}     & 79.7     \\
{[}14,14{]}, {[}14,14{]}, ..., {[}14,14{]}, {[}14,14{]} & 78.8    \\
{[}7,7{]}, {[}14,14{]}, ..., {[}7,7{]}, {[}14,14{]}   & 80.0     \\
{[}7,14{]}, {[}7,14{]}, ..., {[}7,14{]}, {[}7,14{]}   & \textbf{80.2}     \\
\bottomrule
\end{tabular}}
\end{center}
\caption{Results of different combinations of patch size on ImageNet classfication. Using multiple patch sizes is a better setting. }
\label{patchsize}
\end{table}

\noindent
\medskip\\
\textbf{Impact of different patch sizes:}
Patch size is an important hyper-parameter in ConTNet. In order to explore the effect of different patch sizes and find a suitable patch size, we test four different arrangements of patch size by tuning ConT-M: $1)$ All patch size is set to 7, $2)$ all patch size is to 14, $3)$ the patch size of the first block in each stage is set to 7 and that of the second block is set to 14, $4)$ the patch size of first STE in each block is set to 7 and that of second STE is set to 14. Table~\ref{patchsize} shows that except setting $2)$, all patches are $14 \times 14$, the results of other arrangements of patch sizes are relatively close. So we choose the patch size of 7 and 14 alternately as the default setting.



\begin{table}[]
\begin{center}
\setlength{\tabcolsep}{5.3mm}{
\begin{tabular}{@{}lccc@{}}
\toprule
Network & Conv LR & STE LR & Top-1(\%) 
\\ 
\midrule
ConT-M   & 0.2     & 0.01   & 77.4     
\\ 
ConT-M   & 0.2     & 0.005  & 77.6     
\\ 
ConT-M   & 0.1     & 0.005  & 77.6     
\\ 
ConT-M   & 0.1     & 0.001  & 77.1     
\\ 
\bottomrule
\end{tabular}}
\end{center}
\caption{Results of different pairs of learning rates on ImageNet validation. The network is trained with SGD optimizer. }
\label{lrrate}
\end{table}

\noindent
\medskip\\
\textbf{Range of learning rate:}
We would also like to know the impacts of different learning rates on our model. A wide range of learning rate choices means that it can be applied to various downstream tasks and transferred stably. We train ConT-M with SGD optimizer and adjust the learning rate of conv layers and STE separately. In our practice, the initial learning rate of STE is denoted by $lr_{ste} \in \{0.01, 0.005, 0.001\}$ and the initial learning rate of conv layer is denoted by $lr_{conv} \in \{0.2,0.1\}$. Table~\ref{lrrate} shows the results of ConT-M trained with each group of learning rates and we find that the performance of ConT-M is not susceptible to learning rate. Therefore, the performance of applying ConTNet as the backbone of computer vision models can not be limited by the specific learning rate setting adopted in other tasks, which suggests that ConTNet is of great robustness.

\noindent
\medskip\\
\textbf{Group convolution:}
ConTNet is equipped with multiple conv layers that aggregate local features supplementary to global information captured by STE. Group convolution is a successful variants of convolution to reduces computational cost. In authentic convolution, the weight of kernel is denoted by $\mathbf{W_{\rm{k}}} \in \mathbb{R} ^ \mathit{k \times k \times C_{in} \times C_{out}}$, which is grouped into a set of kernels ${ \mathbf{W_{\rm{1}}}, \mathbf{W_{\rm{2}}},...,\mathbf{W_{\rm{g}}} \in \mathbb{R} ^ {\mathit{k \times k \times C_{in}/g \times C_{out}/g }}  }$ in a group convolution. The grouped convolutional kernel disentangles the channel number and therefore achieve a lower computational and memory costs than authentic convolution. Table~\ref{group} shows the impact of different group settings on performance of ConT-M, which reveals that grouped convolution 
induces a non-negligible performance degradation for ConTNet. An alternative solution to increase efficiency of our model is to exploit depth-wise separable convolution. As can be seen from Table~\ref{group}, depth-wise separable convolution make the best trade-off between the computational cost and inference performance.

\begin{table}[]
\begin{center}
\setlength{\tabcolsep}{3mm}{
\begin{tabular}{@{}lccc@{}}
\toprule
Group number & FLOPs(G) & \#Param(M) & Top-1(\%) 
\\ 
\midrule
1   & 3.1     & 19.2  & 80.2     
\\ 
4   & 2.6     & 15.0  & 78.8     
\\ 
8   & 2.5     & 14.3  & 79.3     
\\ 
16   & 2.4    & 14.0  & 79.1     
\\ 

depthwise & 2.5& 15.3 & 79.6
\\

\bottomrule
\end{tabular}}
\end{center}
\caption{Results of replacing convolution in ConT Block with group convolution with different number of groups. "depthwise" means that the convolution is depth-wise separable. }
\label{group}
\end{table}

\subsection{Visualization}

\begin{figure*}[t]
\begin{center}
   \includegraphics[width=1.0\textwidth]{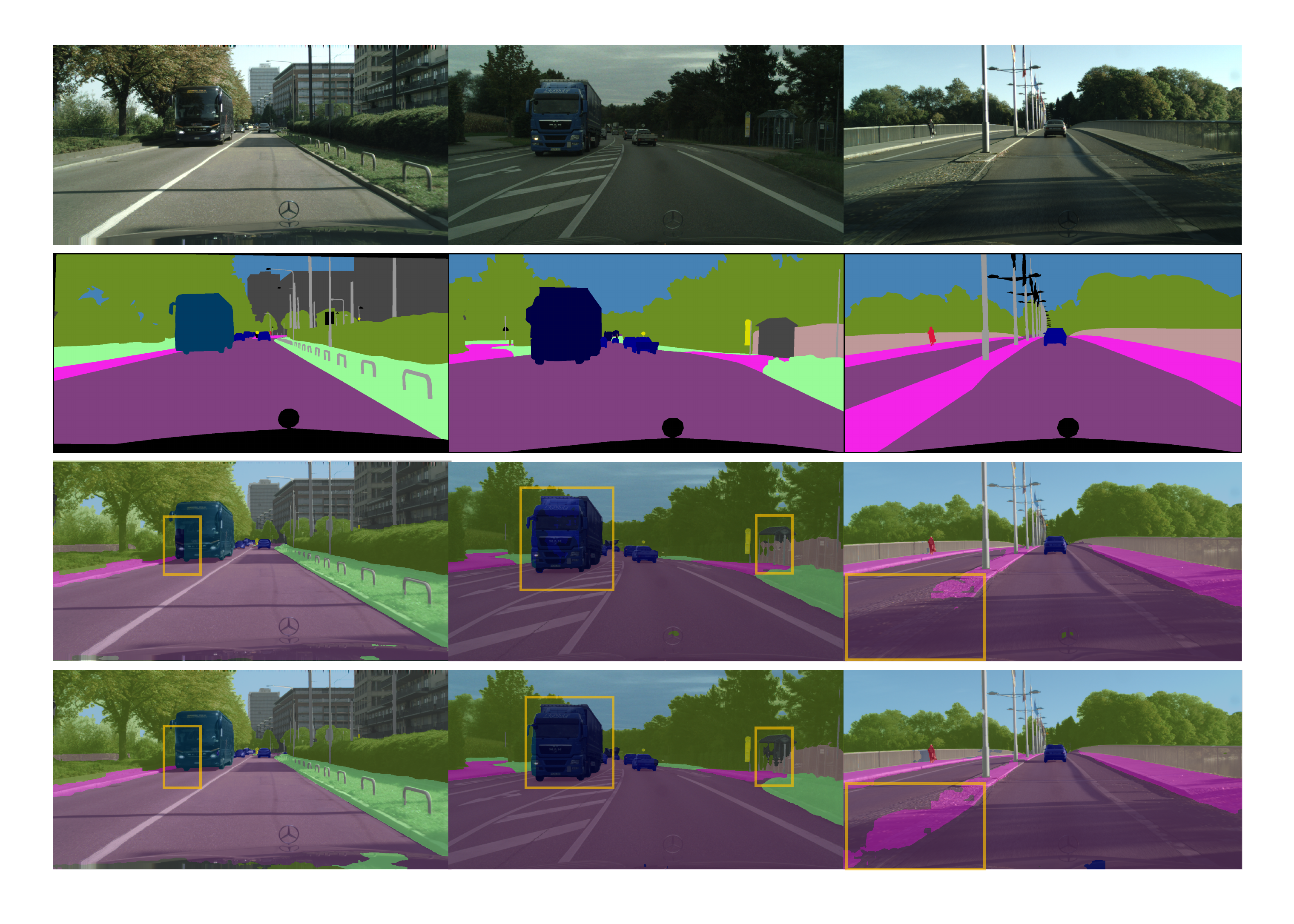}
\end{center}
\caption{Visualized comparison of segmentation results for PSPNet on Cityscapes validation set. Each column is a group of sample from validation set and the top two rows are the raw image with its corresponding groud truth. The bottom two rows represents the results of PSPNet with ResNet50 backbone and ConT-M backbone separately. The yellow bounding box marks the region where the segmentation results presents a considerable variance.}
\label{fig: vis_seg}
\end{figure*}

\noindent
\textbf{Visualization of segmentation:} We list some visualized results of semantic segmentation to compare performance qualitatively as shown in Figure~\ref{fig: vis_seg}. In the first column, the true class of the area inside the yellow box is labeled as bus both in ground truth and the fourth row. However, in the third row, these pixels are labeled as truck. In the second column, inside the left yellow box is a truck as shown in the second and fourth tow, but in the third row, some pixels of the truck are labeled as car. As for the right yellow box in the second column, the results of the fourth row labeled more pixels than that of the third row. In the third column, a large area of side walk is left unlabeled due to the occlusion of shadow as shown in the yellow box. In contrast, as can be seen from the fourth row, almost all of the sidewalk pixels that were missed in the third row are labeled.

The segmentation results indicated that ConTNet aggregates more global information than ConvNets and the pixels of a large object enjoy a better inner consistency. Such a promotion of segmentation network is stemmed from the ability of modeling long-range dependencies by exploiting STE in ConTNet.

\section{Concolusion}
In this work, we propose an innovative convolution-transformer network (ConTNet). ConTNet combines transformer and convolution by stacking standard transformer encoder (STE) and spatial convolution alternately. We introduce a weight-shared patch-wise STE to model a large kernel. A series of experiments demonstrate that ConTNet outperforms ResNet and achieves comparable accuracy of trendy visual transformer with much lower computational complexity. We verify that the effectiveness of implementing downstream tasks with ConTNet backbone is primarily derived from aggregating more contextual information. Moreover, we demonstrate that ConTNet is as robust as ConNets for example ResNet. In other words, ConTNet's performance does not rely on strong data augmentations and fancy training tricks, and is not sensitive to hyper-parameters as well. Finally, in ConTNet we use the vanilla convolution and transformer, if replaced with the improved versions of convolution and transformer recently, a further performance is promised. We hope that our work could bring some new ideas for model design. 

{\small
\bibliographystyle{ieee}
\bibliography{egbib}
}

\end{document}